\documentclass[a4paper,fleqn]{cas-dc}

\usepackage[numbers]{natbib}
\makeatletter
\renewcommand\citet{%
  \@ifnextchar[{\NAT@citet}{\NAT@citet[]}%
}
\def\NAT@citet[#1]#2{%
  \citeauthor{#2} \citep[#1]{#2}%
}
\makeatother

\usepackage{amssymb}
\usepackage{amsmath}
\usepackage{array}
\usepackage{enumerate}
\usepackage{enumitem}
\usepackage{csquotes}
\usepackage[english]{babel}
\usepackage{multicol}
\usepackage{multirow}
\usepackage{xcolor}
\usepackage{subcaption}
\usepackage{stmaryrd}
\usepackage{algorithm}
\usepackage{algpseudocode}
\usepackage{cancel}
\usepackage{makecell}
\usepackage{float}
\usepackage{tabularx}
\usepackage{graphics}
\usepackage{graphicx}
\usepackage{adjustbox}

\newtheorem{definition}{Definition}[section]

\usepackage[acronym]{glossaries}
\newacronym{flops}{flops}{floating-point operations}
\newacronym{pd}{pd}{positive definite}
\newacronym{spd}{spd}{symetric positive definite}
\newacronym{wmi}{WMI}{Woodbury Matrix Identity}
\newacronym{sm}{SM}{Sherman-Morrison}
\newacronym{ism}{ISM}{Iterative Sherman-Morrison}
\newacronym{di}{DI}{Direct Inversion}
\newacronym{cf}{CF}{Christoffel function}
\newacronym{cdk}{CD-Kernel}{Christoffel-Darboux Kernel}
\newacronym{sos}{SOS}{sum-of-squares}

\usepackage{tikz}
\usetikzlibrary{matrix, positioning, fit, backgrounds}

\newcommand{\R}[0]{\mathbb{R}}
\newcommand{\N}[0]{\mathbb{N}}
\newcommand{\veca}[0]{\mathbf{a}}
\newcommand{\vecb}[0]{\mathbf{b}}
\newcommand{\vecp}[0]{\mathbf{p}}

\newcommand{\vecw}[0]{\mathbf{w}}
\newcommand{\vecx}[0]{\mathbf{x}}
\newcommand{\vecy}[0]{\mathbf{y}}
\newcommand{\vecz}[0]{\mathbf{z}}
\newcommand{\vecalph}[0]{\mathbf{\alpha}}

\newcommand{\mysum}[2]{\underset{#1}{\overset{#2}{\sum}}}

\begin{document}
\let\WriteBookmarks\relax
\def\floatpagepagefraction{1}
\def\textpagefraction{.001}
\shorttitle{Cost Trade-offs in Matrix Inversion Updates for Streaming Outlier Detection}

\shortauthors{Grivet and Trav\'e-Massuy\`es}


\title[mode=title]{Cost Trade-offs in Matrix Inversion Updates for Streaming Outlier Detection}

\author[1,2]{Florian Grivet}[type=editor, orcid=0009-0007-7096-3258]
\cormark[1]
\ead{florian.grivet@cnes.fr}

\credit{Writing - original draft, Writing - review and editing, Conceptualization, Formal analysis, Methodology, Software, Data curation, Investigation, Validation}

\affiliation[1]{organization={CNES},
                addressline={18 Av. Edouard Belin}, 
                city={Toulouse},
                postcode={31400}, 
                city={Toulouse},
                country={France}}

\author[2]{Louise Trav\'e-Massuy\`es}[orcid=0000-0002-5322-8418]
\fnmark[1]
\ead{louise@laas.fr}

\credit{Writing - review and editing, Supervision, Funding acquisition, Resources, Conceptualization, Validation}

\affiliation[2]{organization={LAAS-CNRS, University of Toulouse, CNRS},
                addressline={7 Av. du Colonel Roche}, 
                postcode={31400}, 
                city={Toulouse},
                country={France}}

\cortext[cor1]{Corresponding author}
\fntext[fn1]{Head of ANITI chair ADDX}

\begin{abstract}
Outlier detection identifies data points that deviate significantly from expected patterns, revealing anomalies that may require special attention. Incorporating online learning further improves accuracy by continuously updating the model to reflect the most recent data. When employing the Christoffel function as an outlier score, online learning requires updating the inverse of a matrix following a rank-$k$ update, given the initial inverse. Surprisingly, there is no consensus on the optimal method for this task. This technical note aims to compare three different updating methods: Direct Inversion (DI), Iterative Sherman-Morrison (ISM), and Woodbury Matrix Identity (WMI), to identify the most suitable approach for different scenarios. We first derive the theoretical computational costs of each method and then validate these findings through comprehensive Python simulations run on a CPU. These results allow us to propose a simple, quantitative, and easy-to-remember rule that can be stated qualitatively as follows: ISM is optimal for rank-1 updates, WMI excels for small updates relative to matrix size, and DI is preferable otherwise. This technical note produces a general result for any problem involving a matrix inversion update. In particular, it contributes to the ongoing development of efficient online outlier detection techniques.
\end{abstract}

\begin{keywords}
Outlier Detection \sep Christoffel function \sep Rank update \sep Matrix Inversion \sep Sherman-Morrison \sep Woodbury Matrix Identity \sep Computational cost
\end{keywords}

\maketitle


\section{Introduction}
\label{sec:intro}

The detection of outliers in data streams has become increasingly important in a wide range of applications, from fraud detection to quality control in manufacturing. In such settings, data arrive sequentially and often at high rates, making online learning approaches particularly attractive. These methods continuously update models as new observations become available, allowing anomaly detection systems to adapt to evolving data distributions while maintaining strong performance over time. 

\vspace{0.5em}

Among recent approaches to anomaly detection in data streams \cite{app13106353, Zhou2025qx}, \citet{article_kevin} introduce an outlier scoring mechanism based on the \gls*{cf} \cite{sos}. This score is defined in terms of the inverse of a \acrlong*{spd} moment matrix associated with the data. In streaming settings, this matrix is updated sequentially via rank-$k$ corrections as new observations arrive. While the resulting CF scores are invariant to the specific inverse update strategy employed -- up to numerical precision -- the choice of update method has a substantial impact on computational cost, numerical stability, and scalability. These considerations are critical in streaming settings, where efficiency directly constrains real-time applicability.

\vspace{0.5em}

Several strategies are available for updating matrix inverses after rank-$k$ corrections, including \gls*{di}, \gls*{ism} \cite{sherman_morrison}, and the \gls*{wmi} \cite{woodbury1950inverting}. Despite their widespread use, there is currently no clear quantitative guidance on which method is preferable under different conditions, such as varying matrix size $s$ or update rank $k$. This lack of guidance can lead to inefficient implementations that unnecessarily limit the practicality of CF-based anomaly detection in streaming environments.

\vspace{0.5em}

This article is presented as a technical note aimed at addressing this gap by comparing inverse update strategies in the context of Christoffel-function-based outlier detection. It does not propose a new scoring model, but instead analyzes how different matrix update methods may affect computational efficiency.

\vspace{0.5em}

The contributions of this technical note are summarized as follows: \vspace{-0.5em}
\begin{itemize}
    \itemsep 0em
    \item This note introduces the Christoffel function, explores its key properties, and presents DyCF, a frugal streaming outlier detection method inspired by these foundations, that motivates the paper's work.
    \item This work derives the computational costs of three matrix inverse rank-k update methods, namely Direct Inversion, Iterative Sherman-Morrison, and Woodbury Matrix Identity. 
    \item Summarizing and comparing the three theoretical computational costs yields a unified reference, and the findings are validated through comprehensive Python simulations run on a CPU. 
    \item As a key takeaway, this note offers a simple, quantitative, and easy-to-remember rule, expressed in terms of the matrix dimension $s$ and the update rank $k$, for selecting among the three rank-$k$ matrix inverse update strategies implemented in Python on CPU.
\end{itemize}

\vspace{0.5em}

The technical note is organized as follows. Section \ref{sec:christoffel_function} briefly reviews the Christoffel function, highlighting the properties relevant to anomaly detection. Section \ref{sec:data_streams} then discusses its use in streaming outlier detection, emphasizing the need for efficient inverse updates under rank-$k$ corrections. Section \ref{sec:computational_costs} presents the Direct Inversion, Iterative Sherman–Morrison, and Woodbury Matrix Identity approaches, detailing their algorithms and theoretical computational costs. Section \ref{sec:cost_comparison} compares the theoretical computational costs of the three methods. Section \ref{sec:optimal_method_selection} analyzes theoretical predictions with empirical results to derive practical implementations. Finally, Section \ref{sec:conclusion} summarizes the main findings, resumes practical guidance, discusses limitations, and points at interesting topics for future work.

\section{The Christoffel function}
\label{sec:christoffel_function}

The \gls*{cf} originates from the theory of approximation and orthogonal polynomials. \citet{sos} demonstrated that the \gls*{cf} is related to a \gls*{sos} polynomial whose sublevel set effectively captures the shape of a dataset. Building on this discovery, \citet{data_analysis_christoffel}, and \citet{livre_christoffel} developed a comprehensive theoretical framework for data analysis, and in particular, anomaly detection.

\vspace{0.5em}

This section explores key properties of both the theoretical \gls*{cf} (referred to as the population \acrlong*{cf}) and its empirical counterpart.

\subsection{The population \acrlong*{cf}}
Let $\vecx = \begin{pmatrix} x_1, x_2, \cdots, x_d \end{pmatrix} \in \R^d$. To define polynomials, we adopt the multi-index notation $\vecalph = \left( \alpha_i \right)_{i=1...d} \in \N^d$, such that the monomial $\vecx^\vecalph$ of total degree $deg(\vecx^{\vecalph}) = |\vecalph| = \sum_{i=1}^{d} \alpha_i$ is given by $\vecx^\vecalph = x_1^{\alpha_1} x_2^{\alpha_2} \cdots x_d^{\alpha_d}$. In short form, we denote the set of $d$-variate polynomials by $\R[\vecx]$. The dimension of $\R_n[\vecx]$, the space of $d$-variate polynomials of degree at most $n$, is given by $s_d(n) = \begin{pmatrix} d+n \\ n\end{pmatrix}$. \\
Let $\{P_i : 1 \leq i \leq s_d(n) \}$ be a basis of $\R_n[\vecx]$. We denote 
\begin{align*}
    v_n : \R^d & \longrightarrow \R^{s_d(n)} \\
    \vecx & \longmapsto \left(P_1(\vecx), P_2(\vecx), \cdots, P_{s_d(n)}(\vecx) \right)^T
\end{align*}
The monomials in $v_n(\vecx)$ are graded in the lexicographic order\footnote{lexicographic order: monomials are first ordered according to ascending total degree $|\vecalph|$, and then using lexicographic order on variables considering $\vecx_1=a, \vecx_2=b$, etc.}.


Let $\Omega \subset \R^d$ be a compact set, with non-empty interior. Let $\mu$ be a Borel measure supported on $\Omega$ and define the associated moment matrix.

\begin{definition}[The moment matrix]
\label{def:moment_matrix}
The moment matrix of degree $n \in \N$, associated with measure $\mu$, denoted by $M_n(\mu) \in \R^{s_d(n) \times s_d(n)}$, is defined as
\begin{equation}
    \label{eq:mm}
    M_n(\mu) = \int_{\R^d} v_n(\vecx) ~ v_n(\vecx)^T d\mu(\vecx).
\end{equation}
\end{definition}

\newpage

Note that this matrix is \acrlong*{spd}, thus non-singular for all $n$ (see \citep[Section 2.2]{sos} or \citep[Remark 2.3]{vu2020rateconvergencegeometricinference} for the proof).

\vspace{0.5em}
The population Christoffel function is defined as follows.
\begin{definition}[The population \acrlong*{cf}]
\label{def:population_christoffel}
The population \acrlong*{cf} of degree $n \in \N$, associated with the measure $\mu$, denoted by $\Lambda^\mu_n(\vecx)$, is defined as
\begin{equation}
    \label{eq:cf_int_p}
    \Lambda^\mu_n(\vecx) = \underset{P \in \R_n[\vecx]}{min}\left\{ \int_\Omega P^2(\vecz) ~ d\mu(\vecz), \quad P(\vecx) = 1 \right\}.
\end{equation}
\end{definition}

Now, for any polynomial $P \in \R_n[\vecx]$, there exists some $\vecp \in \R^{s_d(n)}$ such that $P(\vecx) = \vecp^T v_n(\vecx)$ for any $\vecx \in \R^d$. Thus, the objective function becomes $\int_\Omega \vecp^T v_n(\vecz) ~ v_n(\vecz)^T \vecp ~ d\mu(\vecz) = \vecp^T M_n(\mu) ~ \vecp$, so that
\begin{equation}
    \label{eq:cf_mm}
    \Lambda^\mu_n(\vecx) = \underset{\vecp \in \R^{s_d(n)}}{min}\left\{ \vecp^T M_n(\mu) ~ \vecp, \quad \vecp^T v_n(\vecx) = 1 \right\}. 
\end{equation}

The \acrlong*{cdk}, which is defined below, is related to the \acrlong*{cf}.
\begin{definition}[The \acrlong*{cdk}]
\label{def:cd_kernel}
The \\ \gls*{cdk} associated with the measure $\mu$, denoted by $K^\mu_n(\vecx, \vecy)$, is defined as
\begin{equation}
    \label{eq:cdk}
    (\vecx, \vecy) \mapsto K^\mu_n(\vecx, \vecy) = v_n(\vecx)^T M_n(\mu) ~ v_n(\vecy),
\end{equation}
\end{definition}
while the polynomial $Q_{\mu, n}$ reads
\begin{equation}
    \label{eq:Q}
    \vecx \mapsto Q_{\mu, n}(\vecx) = K^\mu_n(\vecx, \vecx) = v_n(\vecx)^T M_n(\mu)^{-1} v_n(\vecx).
\end{equation}
$Q_{\mu, n}$ is a \acrfull*{sos} polynomial of degree $2n$. An interesting property of this \gls*{sos} polynomial is its behavior inside and outside its support $\Omega$. \citet[Lemma 4.3.1]{livre_christoffel} quantifies at least the exponential growth with $n$ for data points outside the support, while inside, it is at most polynomial \citep[Lemma 4.3.2]{livre_christoffel}. The \gls*{cf} of degree $n \in \N$ defined in equation \eqref{eq:cf_int_p} can be rewritten as 
\begin{equation}
    \Lambda^\mu_n(\vecx) = \frac{1}{Q_{\mu, n}(\vecx)} = \frac{1}{v_n(\vecx)^T M_n(\mu)^{-1} v_n(\vecx)}
\end{equation}
and we have
\begin{equation}
\Lambda^\mu_n(\vecx)^{-1} = Q_{\mu, n}(\vecx),
\end{equation}
so that $\Lambda^\mu_n(\vecx)^{-1}$ inherits from the properties of $Q_{\mu, n}(\vecx)$.

\subsection{The empirical \acrlong*{cf}}

In practical applications, the measure $\mu$ is unknown. Let $\mathcal{X}$ be a cloud of $N$ data points $\vecx \in \R^d$ sampled from the theoretical measure $\mu$ supported on $\Omega$. We define the discrete measure $\mu_N$ supported on $\mathcal{X}$ such that $\mu_N = \frac{1}{N} \sum_{\vecx \in \mathcal{X}} \delta_{\vecx}$ where $\delta_\vecx$ corresponds to the Dirac measure at $\vecx$. The empirical version of the moment matrix can be written as
\begin{equation}
    M_n(\mu_N) = \frac{1}{N} \mysum{\vecx \in \mathcal{X}}{} v_n(\vecx) ~ v_n(\vecx)^T.
    \label{eq:empirical_mm}
\end{equation}
\citet[Corollary 6.3.5]{livre_christoffel} guarantees that the matrix $M_n(\mu_N)$ is invertible if the size of $\mathcal{X}$: $N$, is greater than $s_d(n)$, e.g. the number of samples is greater than the size of the moment matrix.

\begin{definition}[The empirical \acrlong*{cf}]
\label{def:empirical_cf}
    Under the condition $|\mathcal{X}| = N > s_d(n)$, the empirical \gls*{cf} is defined as
    \begin{equation}
        \label{eq:empirical_cf}
        \Lambda_n^{\mu_N}(\vecx) = \frac{1}{v_n(\vecx)^T M_n(\mu_N)^{-1} ~ v_n(\vecx)}.    
    \end{equation}
\end{definition}

According to \citet[Theorem 3.13]{data_analysis_christoffel}, the empirical \gls*{cf} converges to the population \gls*{cf} as $N$ increases:
\vspace{-0.2em}
\begin{equation*}
    \| \Lambda_n^{\mu_N} - \Lambda_n^{\mu} \|_\infty = \underset{\vecx \in \R^d}{sup} \left \{|\Lambda_n^{\mu_N}(\vecx) - \Lambda_n^{\mu}(\vecx)| \right \} \underset{N \to \infty}{\longrightarrow} 0 \quad a.s.
\end{equation*}

\section{The \acrlong*{cf} for outlier detection in data streams}
\label{sec:data_streams}
\citet{article_kevin} introduce DyCF, a novel outlier detection algorithm for data streams leveraging the \acrlong*{cf}. This algorithm supports online learning using a rank-$k$ update. This section details the anomaly detection method based on the \gls*{cf} as well as the online learning principle of the method.

\subsection{The Christoffel function for outlier detection}
As mentioned in \autoref{sec:christoffel_function}, the polynomial $Q_{\mu, n}(\vecx)$, and hence $\Lambda_n^{\mu}(\vecx)^{-1}$, effectively captures the shape of the underlying dataset. Furthermore, as explained above, there exists a dichotomy in the growth behavior of $\Lambda_n^{\mu}(\vecx)^{-1}$: it exhibits at most polynomial growth when $\vecx$ is within the support $\Omega$ and at least exponential growth when $\vecx$ is outside $\Omega$. Since $\Lambda_n^{\mu_N}(\vecx)^{-1}$ converges to $\Lambda_n^{\mu}(\vecx)^{-1}$, these properties are preserved for finite datasets. Consequently, $\Lambda^{\mu_N}_n(\vecx)^{-1}$ is well-suited as a scoring function for outlier detection. One can define a level set, or threshold, $\gamma_{n, d}$ such that all points $\vecx \in \R^d$ with a value of $\Lambda^{\mu_N}_n(\vecx)^{-1}$ higher than $\gamma_{n,d}$ are considered as outliers. This define the scoring function
\begin{equation}
    \label{eq:scoring_function}
    S_{n,d}(\vecx) = \frac{\Lambda^{\mu_N}_n(\vecx)^{-1}}{\gamma_{n,d}},
\end{equation}
where a point is detected as an outlier if $S_{n,d} \geq 1$.

\subsection{Online learning in DyCF}
Considering $\mathcal{X}$ as a dataset, $M_n(\mu_N)$ can be seen as a summary or an encoding of this dataset. In a data stream, if a new nominal instance $\vecx$ arrives, we can improve the performance of the algorithm by integrating this point into the database $\mathcal{X}$ and updating the moment matrix $M_n(\mu_N)$. This is called online learning. The first update method would be to recalculate the moment matrix $M_n(\mu_N)$ using equation \eqref{eq:empirical_mm} and to invert it. However, this last operation is very costly. Nevertheless, we can notice that: 
\[
\begin{aligned}
M_n(\mu_{N+1})
    &= \frac{1}{N+1} 
       \mysum{\vecz \in \mathcal{X} \cup \{\vecx\}}{} 
       v_n(\vecz)\, v_n(\vecz)^T \\[0.3em]
    &= \frac{1}{N+1} \left( 
       N\, M_n(\mu_N) + v_n(\vecx)\, v_n(\vecx)^T 
       \right).
\end{aligned}
\]
Thus, we can use the \acrlong*{sm}'s formula or the \acrlong*{wmi} to avoid recalculating the inverse of $M_n(\mu_{N+1})$, which is an $s_d(n) \times s_d(n)$ matrix. 

\vspace{0.5em}

When the update involves $k \in \N^*$ new data points $\vecx \in \R^d$, the online learning phase goes as follows in three steps:
\begin{enumerate}[label=(\roman*)]
    \item We denormalize $M_n(\mu_N)$ to obtain either \\ $M = M_{n}(\mu_N) \times N$, or $M^{-1} = M_{n}(\mu_N)^{-1} / N$.
    \item The \gls*{di} method calculates $M^{-1}_{updated}$ from $M$ using
    \begin{equation}
        \label{eq:problem_m_updated}
        M^{-1}_{updated} = \left(M + \mysum{i=1}{k}v_n\left(\vecx^{(i)}\right) v_n\left(\vecx^{(i)}\right)^T\right)^{-1}
    \end{equation}
    and the \gls*{ism} and \gls*{wmi} methods calculate $M^{-1}_{updated}$ from $M^{-1}$ using \gls*{sm} and \gls*{wmi} formulas, e.g. equations \eqref{eq:sherman_problem} and \eqref{eq:woodbury} given below, respectively.
    \item Renormalize $M^{-1}_{updated}$ to obtain the updated inverse moment matrix $M_n(\mu_{N+k})^{-1} = M^{-1}_{updated} \times (N + k)$.
\end{enumerate}

\paragraph{The Sherman-Morrison formula and the Woodbury Matrix Identity --}
Suppose $A \in \mathbb{R}^{n \times n}$ is an invertible square matrix and $u, v \in \mathbb{R}^n$ are column vectors. Then $A + u v^T$ is invertible if and only if $1 + v^T A^{-1} u \neq 0$. In this case, the SM formula \citep{sherman} states the following
\begin{equation}
    (A + uv^T)^{-1} = A^{-1} - \frac{A^{-1} u v^T A^{-1}}{1 + v^T A^{-1} u},
    \label{eq:sherman}
\end{equation}

\noindent which, in our case, becomes:
\begin{equation}
    \label{eq:sherman_problem}
    \resizebox{0.9\linewidth}{!}{$
        \left(M + v_n(\vecx) v_n(\vecx)^T\right)^{-1} = M^{-1} - \frac{M^{-1} v_n(\vecx) v_n(\vecx)^T M^{-1}}{1 + v_n(\vecx)^T M^{-1} v_n(\vecx)}.
    $}
\end{equation}

Now, let us recall the \acrfull*{wmi}:
\vspace{-0.5em}
\begin{equation}
    \resizebox{0.9\linewidth}{!}{$
        \left(A + UCV\right)^{-1} = A^{-1} - A^{-1}U\left(C^{-1} + VA^{-1}U\right)^{-1}VA^{-1},
    $}
    \label{eq:woodbury}
\end{equation}

\noindent which, in our case, becomes:
\begin{equation}
    \resizebox{0.9\linewidth}{!}{$
        \left(M + X^TIX\right)^{-1} = M^{-1} - M^{-1}X^T\left(I^{-1} + XM^{-1}X^T\right)^{-1}XM^{-1},
    $}
    \label{eq:ourwoodbury}
\end{equation}
with $I$ the identity matrix of size $k$ and $X$ the design matrix: $X = \begin{bmatrix} v_n\left(\vecx^{(1)}\right) & v_n\left(\vecx^{(2)}\right) & \cdots & v_n\left(\vecx^{(k)}\right) \end{bmatrix}^T \in \R^{k \times s}$.

\vspace{1em}

Note that the normalization costs of steps (i) and (iii) are the same regardless of the method used. The most efficient method will therefore be the one with the lowest cost in step (ii). Moreover, the \gls*{di} method updates both $M_{n}(\mu_N)$ and $M_{n}(\mu_N)^{-1}$. However, to detect outliers, since $\Lambda_n^{\mu_N}$ only uses $M_{n}(\mu_N)^{-1}$, we do not need to compute $M_{n}(\mu_{N+k})$, so we can use the \gls*{ism} or \gls*{wmi} methods.

\section{Computational costs}
\label{sec:computational_costs}
For the sake of simplifying calculations, the size of the moment matrix $s_d(n)$ will be referred to as $s$ in this section. \autoref{sec:computation_cost_generic} reports the costs of the intermediate steps used in this section in terms of \gls*{flops}.


\subsection{Computational cost of the DI method for a rank-$k$ update}
\label{sec:inverse}

In this subsection, we calculate the computational cost of the \gls*{di} method and provide an algorithm for it.

\subsubsection{Computational cost of a matrix inversion}
The computational cost of inverting the moment matrix $M_n(\mu_N)$ of size $s \times s$ is dependent on the algorithm employed. Using an LU factorization, the computational cost of $M_n(\mu_N)^{-1}$ is expressed as \citep[Theorem 2.31]{matrixdecomposition}:
\begin{equation}
    O\left(2 s^3\right) ~ \gls*{flops}.
    \label{eq:non_symmetric_inversion}
\end{equation}

Since our moment matrix $M_n(\mu_N)$ is \gls*{spd}, employing Cholesky decomposition reduces the cost to \citep[Section III.A]{cholesky}:
\begin{equation}
    O\left(\frac{5}{6}s^3\right) ~ \gls*{flops}.
    \label{eq:symmetric_inversion}
\end{equation}

\subsubsection{Rank-$k$ update computational cost}
To perform a rank-$k$ update and apply equation \eqref{eq:problem_m_updated}, we first need to compute $\mysum{i=1}{k}v_n\left(\vecx^{(i)}\right) v_n\left(\vecx^{(i)}\right)^T$. This involves performing $k$ column-vector by row-vector products as described in equation \eqref{eq:produit_vecteur_c_vecteur_l}, with a computational cost of $k \times s^2 ~ \gls*{flops}$. Then, we need to sum the $k$ resulting matrices of size $s \times s$, requiring $k-1$ term-by-term matrix additions with a total cost of $(k-1) s^2 ~ \gls*{flops}$. Next, we perform a term-by-term matrix addition with $M$, which incurs a cost of $s^2 ~ \gls*{flops}$.

\vspace{0.5em}

Thus, the computational cost of updating $M$ to form $M_{updated} = M + \sum_{i=1}^{k}v_n\left(\vecx^{(i)}\right) v_n\left(\vecx^{(i)}\right)^T$ amounts to: \\ $k s^2 + (k-1)s^2 + s^2 = 2 k s^2 ~ \gls*{flops}$.

\vspace{0.5em}

Note that if we compute $M_{updated}$ using the design matrix $X = \begin{bmatrix} v_n\left(\vecx^{(1)}\right) & v_n\left(\vecx^{(2)}\right) & \cdots & v_n\left(\vecx^{(k)}\right) \end{bmatrix}^T \in \mathbb{R}^{k \times s}$ and the relation $M_{updated} = M + X^TX$, the computational cost remains equivalent:
1 matrix-by-matrix product according to equation \eqref{eq:produit_matrice_matrice} and 1 term-by-term addition yields a cost of: $2k s^2 - s^2 + s^2 = 2k s^2 ~ \gls*{flops}$. However, considering the superior optimization in Python, we will employ this computational approach for efficiency during our tests.

\vspace{0.5em}

Finally, we need to compute the inverse of $M_{updated}$, which is a \gls*{spd} matrix with the cost described in equation \eqref{eq:symmetric_inversion}.

\vspace{0.5em}

Thus, the computational cost for the \gls*{di} method (\autoref{alg:direct_inversion}) is
\begin{equation}
    O\left(\frac{5}{6} s^3\right) + 2 k s^2 ~ \gls*{flops}.
    \label{eq:direct_inversion_cost}
\end{equation}

\vspace{-0.5em}

\begin{algorithm}[H]
    \caption{\acrlong*{di} Algorithm with $O\left(\frac{5}{6}s^3\right) + 2 ks^2 ~ \gls*{flops}$ \eqref{eq:direct_inversion_cost}}
    \begin{algorithmic}[1]
        \Require Matrix $M$ with size $s \times s$, and vectors $v_n\left(x^{(i)}\right), i = 1 ... k$ to add to $M$;
        \State Construct the design matrix
        \Statex $X = \begin{bmatrix} v_n\left(\vecx^{(1)}\right) & v_n\left(\vecx^{(2)}\right) & \cdots & v_n\left(\vecx^{(k)}\right) \end{bmatrix}^T \in \R^{s \times s}$;
        \State $M_{updated} \gets M + X^TX;$ \hfill $\triangleright ~ 2ks^2 ~ \gls*{flops}$
        \State Compute $M^{-1}_{updated}$ using Cholesky decomposition; 
        \Statex \hfill $ \triangleright ~ O\left(\frac{5}{6}s^3\right) ~ \gls*{flops}$
        \State \textbf{Output} $M_{updated}^{-1};$
    \end{algorithmic}
    \label{alg:direct_inversion}
\end{algorithm}


\subsection{Computational cost of the ISM method for a rank-N update}
\label{sec:sherman}

In this subsection, we calculate the computational cost of the \gls*{ism} method and provide an algorithm for it.

\subsubsection{Computational cost of the numerator}

There are three ways to compute the numerator of the \gls*{sm}'s formula given by equation \eqref{eq:sherman_problem} and recalled below:
\begin{equation*}
    \left(M + v_n(\vecx) v_n(\vecx)^T\right)^{-1} = M^{-1} - \frac{M^{-1} v_n(\vecx) v_n(\vecx)^T M^{-1}}{1 + v_n(\vecx)^T M^{-1} v_n(\vecx)}.
\end{equation*}

\begin{itemize}
    \item \textbf{Compute the outer product first}, i.e., \\ $M^{-1} \left(v_n(\vecx) v_n(\vecx)^T \right) M^{-1}$. This involves 1 column-vector by row-vector product (equation \eqref{eq:produit_vecteur_c_vecteur_l}) and 2 matrix products (equation \eqref{eq:produit_matrice_matrice}), totaling: $s^2 + 2 (2s^3 - s^2) = 4s^3 - s^2 ~ \gls*{flops}$.
    
    \item \textbf{Compute left to right}. This involves 1 product matrix by column-vector (equation \eqref{eq:produit_matrice_vecteur_c}), 1 column-vector by row-vector product (equation \eqref{eq:produit_vecteur_c_vecteur_l}), and 1 matrix product (equation \eqref{eq:produit_matrice_matrice}), totaling: $2s^2 - s + s^2 + 2s^3 - s^2 = 2s^3 + 2s^2 - n ~ \gls*{flops}$.
    
    \item \textbf{Compute the matrices-vectors products first}, i.e., $\left(M^{-1} v_n(\vecx)\right) \left(v_n(\vecx)^T M^{-1}\right)$. This involves 2 matrix-vector products (equations \eqref{eq:produit_matrice_vecteur_c} and \eqref{eq:produit_vecteur_l_matrice}), and 1 column-vector by row-vector product (equation \eqref{eq:produit_vecteur_c_vecteur_l}), totaling: $2(2s^2 - s) + s^2 = 5s^2 - 2s ~ \gls*{flops}$. Moreover, since $M$ is symmetric, $M^{-1} v_n(\vecx) = \left(v_n(\vecx)^T M^{-1}\right)^T$, so we only have 1 matrix-vector product to compute. Thus, the computational cost is $3s^2 - s ~ \gls*{flops}$.
\end{itemize}

The most effective way to compute the numerator of equation \eqref{eq:sherman_problem} is to first perform the matrix-vector products, resulting in a computational cost of $5s^2 - 2s ~ \gls*{flops}$, and in our \gls*{spd} case, a cost of 
\begin{equation}
    3s^2 - s ~ \gls*{flops}.
    \label{eq:numerateur_sherman}
\end{equation}

\subsubsection{Computational cost of the denominator}
Since we have already computed $M^{-1} v_n(\vecx)$ during the numerator computation, we only need to perform 1 row-vector by column-vector product (equation \eqref{eq:produit_vecteur_l_vecteur_c}), and 1 addition which amounts to $2 s - 1 + 1$ \gls*{flops}, i.e.,
\begin{equation}
    2s ~ \gls*{flops}.
    \label{eq:denominateur_sherman}
\end{equation}

\subsubsection{Sherman-Morrison computational cost}
The numerator of equation \eqref{eq:sherman_problem} gives us an $s \times s$ matrix, and the denominator is real. Thus, there is a term-by-term division that costs $s^2 ~ \gls*{flops}$. However, if we perform the division after computing $M^{-1}v_n(\vecx)$ and before the column-vector by row-vector product of the numerator, we only divide a vector of size $s$, reducing the division cost from $s^2$ to $s$.
Finally, we have a term-by-term matrix subtraction that costs $s^2 ~ \gls*{flops}$.
Thus, the computational cost of the \gls*{sm}'s formula for a \gls*{spd} matrix is $\eqref{eq:numerateur_sherman} + \eqref{eq:denominateur_sherman} + s + s^2 = 3s^2 - s + 2s + s + s^2$ \gls*{flops}, i.e.,
\begin{equation}
    4s^2 + 2s ~ \gls*{flops}.
    \label{eq:symmetric_sherman_cost}
\end{equation}

\subsubsection{Iterative Sherman-Morrison computational cost}
The \gls*{sm} formula provides a method for computing the inverse of a matrix that has been modified by a rank-1 update. Specifically, if the matrix is updated with $k$ outer products of the form $v_n(\vecx) v_n(\vecx)^T$, the inverse can be obtained by applying the \gls*{sm} formula iteratively $k$ times. As a result, the computational cost with a rank-$k$ update using the \gls*{ism} method for a \gls*{spd} matrix (\autoref{alg:iterative_sherman_morrison}) is
\begin{equation}
    4ks^2 + 2ks ~ \gls*{flops}.
    \label{eq:N_sherman_cost}
\end{equation}

\vspace{-0.5em}

\begin{algorithm}[H]
    \caption{\acrlong*{ism} Algorithm with $4ks^2 + 2ks ~ \gls*{flops}$ \eqref{eq:N_sherman_cost}}
    \begin{algorithmic}[1]
        \Require The inverse of the matrix $M$: $M^{-1}$ with size $s \times s$, and vectors $\left(v_n\left(x^{(i)}\right)\right)_{i = 1 ... k} \in \R^{s}$ to add to $M$;
        \State Initialize $M_{updated}^{-1} \gets M^{-1}$;
        \For{$i = 1$ to $N$}
            \State {$l \gets M_{updated}^{-1} \times v_n\left(x^{(i)}\right);$} \hfill $\triangleright ~ 2s^2-s~\gls*{flops}$
            \State {$d \gets 1 + v_n\left(x^{(i)}\right)^T \times l;$} \hfill $\triangleright ~ 2s~\gls*{flops}$
            \State {$l_d \gets l / d;$} \hfill $\triangleright ~ s~\gls*{flops}$
            \State {$M_{updated}^{-1} \gets M_{updated}^{-1} - l_d \times l^T;$} \hfill $\triangleright ~ 2s^2 ~\gls*{flops}$
        \EndFor
        \State {\textbf{Output} $M_{updated}^{-1};$}
    \end{algorithmic}
    \label{alg:iterative_sherman_morrison}
\end{algorithm}

\subsection{Computational cost of the WMI method for a rank-N update}
\label{sec:woodbury}

In this subsection, we calculate the computational cost of the \gls*{wmi} method and provide an algorithm for it. The \gls*{wmi} is given by equation \eqref{eq:ourwoodbury} is recalled below:
\begin{equation*}
    \resizebox{0.99\linewidth}{!}{$
        \left(M + X^TIX\right)^{-1} = M^{-1} - M^{-1}X^T\left(I^{-1} + XM^{-1}X^T\right)^{-1}XM^{-1}.
    $}
\end{equation*}

The first step is to compute the matrix products $X M^{-1} X^T$, with a computational cost of equation \eqref{eq:produit_matrice_matrice}: $C^{k \times s}_{s \times s} + C^{k \times s}_{s \times k} =$ \\ $\underbrace{2ks^2 - ks}_{R = X \times M^{-1}} + \underbrace{2k^2s - k^2}_{R \times X^T} = 2 k s^2 + (2k^2 - k) s - k^2 ~ \gls*{flops}$.

\vspace{0.5em}

Next, we perform a term-by-term addition with $I$, which has a computational cost of $k^2 ~ \gls*{flops}$.

\vspace{0.5em}

Then, we need to invert the resulting summed matrix, which by construction is a \gls*{spd} $k \times k$ matrix. The computational cost of this inversion step is $O\left(\frac{5}{6} k^3\right) ~ \gls*{flops}$ \eqref{eq:symmetric_inversion}.

\vspace{0.5em}

Since we have already computed $R = X \times M^{-1}$, and given that $M$ is \gls*{spd}, then $M^{-1} \times X^T = (X \times M^{-1})^T = R^T$. Thus, we only have 2 matrix products to compute, that is \\ $\underbrace{2k^2s - ks}_{Q = R^T \times inv} + \underbrace{2ks^2 - s^2}_{Q \times R} = (2k - 1) s^2 + (2k^2 - k) n ~ \gls*{flops}$.

\vspace{0.5em}

Finally, a term-by-term subtraction is performed, which costs $s^2 ~ \gls*{flops}$.

\vspace{0.5em}

Combining these steps, the overall computational cost for applying the \gls*{wmi} method (\autoref{alg:woodbury_matrix_identity}) is:
\vspace{-0.5em}
\begin{align}
    & 2ks^2 + (2k^2 - k)s - k^2 + k^2 + O\left(\frac{5}{6}k^3\right)  \nonumber \\
    &+ (2k - 1) s^2 + (2k^2 - k) s + s^2 \nonumber \\
    & = 4k s^2 + (4k^2 - 2k) s + O\left(\frac{5}{6}k^3\right) ~ \gls*{flops}
    \label{eq:woodbury_cost}
\end{align}

\vspace{-1em}

\begin{algorithm}[H]
    \caption{\acrlong*{wmi} Algorithm with $4k s^2 + (4k^2 - 2k) s + O\left(\frac{5}{6}k^3\right) ~ \gls*{flops}$ \eqref{eq:woodbury_cost}}
    \begin{algorithmic}[1]
        \Require The inverse of the matrix $M$: $M^{-1}$ with size $s \times s$, and vectors $\left(v_n\left(x^{(i)}\right)\right)_{i = 1 ... k} \in \R^{s}$ to add to $M$;
        \State Initialize $I \in \R^{k \times k}$ and construct the design matrix
        \Statex $X = \begin{bmatrix} v_n\left(\vecx^{(1)}\right) & v_n\left(\vecx^{(2)}\right) & \cdots & v_n\left(\vecx^{(k)}\right) \end{bmatrix}^T \in \R^{s \times s}$;
        \State {$R \gets X M^{-1};$} \hfill $\triangleright ~ 2ks^2 - ks ~ \gls*{flops}$
        \State {$S \gets I + RX^T;$} \hfill $\triangleright ~ 2k^2s ~ \gls*{flops}$
        \State {Compute $S^{-1}$ using Cholesky decomposition;}
        \Statex \hfill $\triangleright ~ O\left(\frac{5}{6}k^3\right) ~ \gls*{flops}$
        \State {$Q \gets R^T S^{-1};$} \hfill $\triangleright ~ 2k^2s - ks ~ \gls*{flops}$
        \State {$M^{-1}_{updated} \gets M^{-1} - QR;$} \hfill $\triangleright ~ 2ks^2 ~ \gls*{flops}$
        \State {\textbf{Output} $M_{updated}^{-1};$}
    \end{algorithmic}
    \label{alg:woodbury_matrix_identity}
\end{algorithm}

\vspace{-1.5em}
\section{Comparison of the computational costs of the DI, ISM, and WMI methods}
\label{sec:cost_comparison}

\begin{table*}[pos=b]
    \centering
    \caption{Computational cost of the three compared methods (in \gls*{flops})}
    \begin{tabular}{c|c|c|c}
         & \makecell{\gls*{di} (\autoref{alg:direct_inversion})} & \makecell{\gls*{ism} (\autoref{alg:iterative_sherman_morrison})} & \makecell{\gls*{wmi} (\autoref{alg:woodbury_matrix_identity})} \\
         \hline
        \makecell{Computational cost} & $O\left(\frac{5}{6} s^3\right) + 2 ks^2$ \eqref{eq:direct_inversion_cost} & $4ks^2 + 2ks$ \eqref{eq:N_sherman_cost} & $4ks^2 + (4k^2 - 2k) s + O\left(\frac{5}{6}k^3\right)$ \eqref{eq:woodbury_cost}
    \end{tabular}
    \label{table:computational_cost}
\end{table*}

We recall that the size of the moment matrix $M_n(\mu_N)$ is $s \times s$, and that we want to add $k$ new data points into $M_n(\mu_N)$.
The computational cost of each method is summarized in \autoref{table:computational_cost}.

The goal of this section is to find the number of new data $k$ to add such that the \gls*{di} cost given by equation \eqref{eq:direct_inversion_cost} is lower than that of the \gls*{ism} method given by equation \eqref{eq:N_sherman_cost} or that of the \gls*{wmi} method given by equation \eqref{eq:woodbury_cost}.

\subsection{Direct inversion method vs. Iterative Sherman-Morrison method}
The condition on $k$ under which the \gls*{di} method is preferable to the \gls*{ism} method is obtained by comparing their computational cost given by  equations \eqref{eq:direct_inversion_cost} and \eqref{eq:N_sherman_cost}, respectively. The condition is obtained by isolating $k$ in the following inequality:
\vspace{-0.2em}
\begin{align}
    & O\left(\frac{5}{6} s^3\right) + 2 k s^2 < 4ks^2 + 2ks \nonumber \\
    & \Longleftrightarrow \quad \frac{5}{6} s^2 < (2s + 2)k \nonumber \\
    & \Longleftrightarrow \quad k > \frac{5s^2}{12\left(s + 1\right)}
    \label{eq:inversion_over_sherman}
\end{align}

\subsection{Direct inversion method vs. Woodbury matrix identity method}
We next compare the \gls*{di} method with the \gls*{wmi} method:
\begin{align}
    & O\left(\frac{5}{6} s^3\right) + 2 ks^2 < 4ks^2 + (4k^2 - 2k) s + O\left(\frac{5}{6}k^3\right) \nonumber \\
    & \Longleftrightarrow \quad \frac{5}{6}s^3 < 2 ks^2 + (4k^2 - 2k)s + \frac{5}{6}k^3 \nonumber \\
    & \Longleftrightarrow \quad \frac{5}{6}k^3 + 4 sk^2 + 2 (s^2 - s) k - \frac{5}{6}s^3 > 0
    \label{eq:inversion_over_woodbury}
\end{align}

Here we have a cubic equation that we need to solve with a fixed s. An empirical approximation is given in \autoref{sec:theoretical_guide}.

\subsection{Woodbury matrix identity method vs. Iterative Sherman-Morrison method}
\label{sec:woodbury_over_sherman}
For $s \geq 1$ and $k \geq 1$, the computation cost of the \gls*{ism} method (equation \eqref{eq:N_sherman_cost}) is generally lower than that of the \gls*{wmi} method (equation \eqref{eq:woodbury_cost}). However, these figures do not account for memory efficiency (read/write operations), nor for the optimization of matrix computations compared to vector calculations and successive iterations, which introduce significant overheads. The measurements carried out in \autoref{sec:experimental_results} show that the \gls*{wmi} method quickly becomes more efficient than the \gls*{ism} method.

\section{Optimal method selection based on the number \texorpdfstring{$k$}{k} of new data}
\label{sec:optimal_method_selection}

\subsection{Theoretical guide for method selection based on new data volume \texorpdfstring{$k$}{k}}
\label{sec:theoretical_guide}

\begin{figure*}[pos=h]
    \centering
    \begin{subfigure}[b]{0.48\textwidth}
        \includegraphics[width=\textwidth]{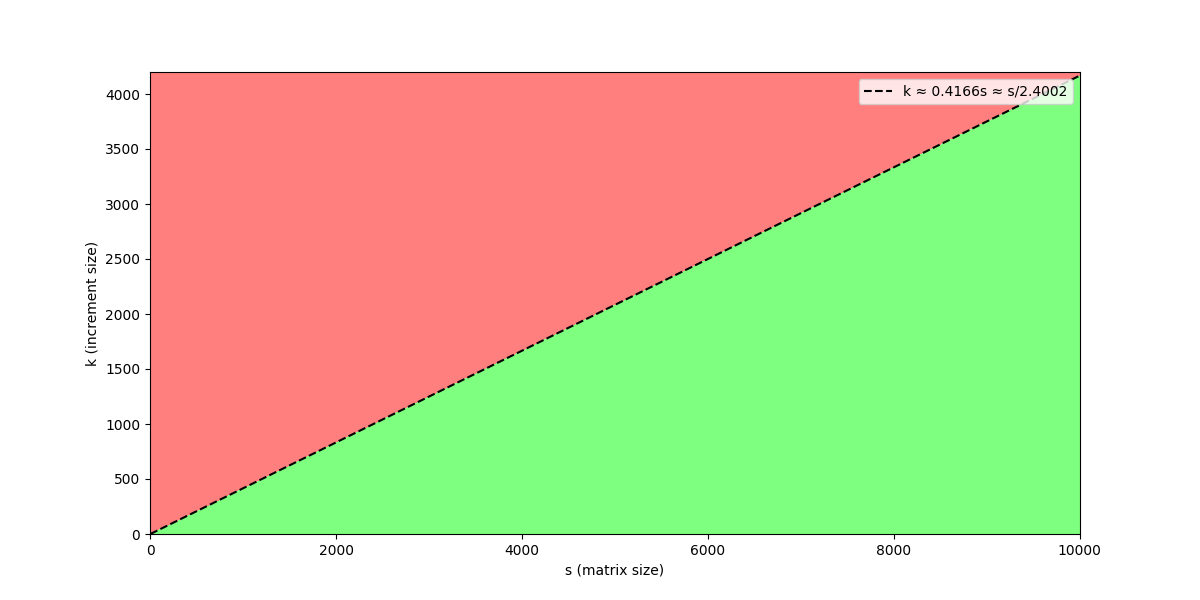}
        \caption{\gls*{di} (\autoref{sec:inverse}) vs. \gls*{ism} (\autoref{sec:sherman})}
        \label{fig:direct_vs_sherman}
    \end{subfigure}
    \hfill
    \begin{subfigure}[b]{0.48\textwidth}
        \includegraphics[width=\textwidth]{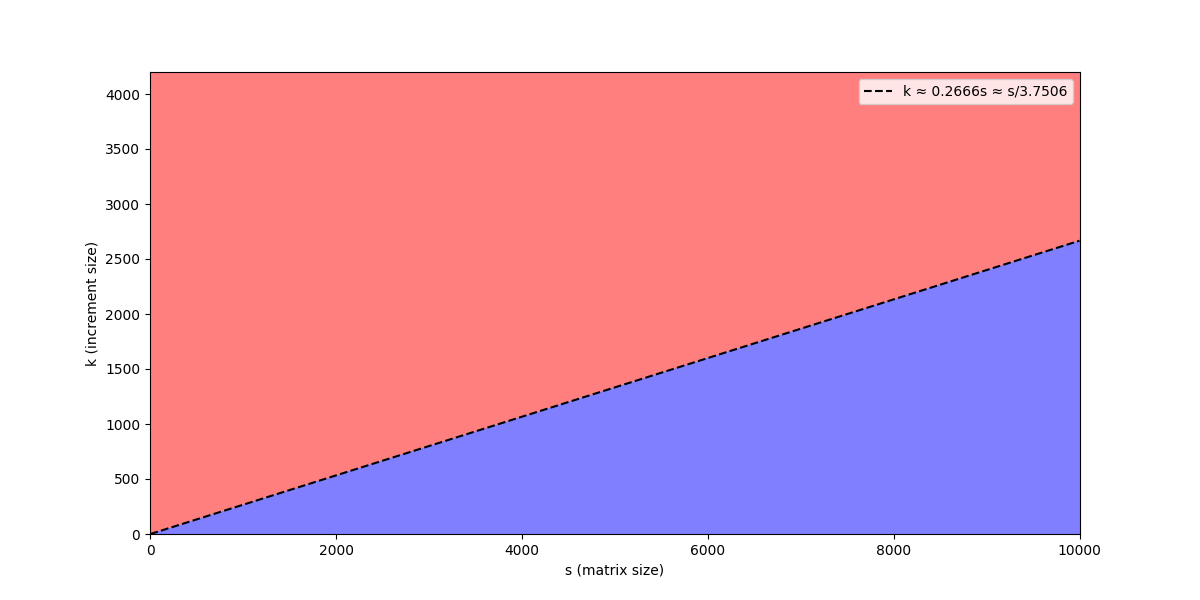}
        \caption{\gls*{di} (\autoref{sec:inverse}) vs. \gls*{wmi} (\autoref{sec:woodbury})}
        \label{fig:direct_vs_woodbury}
    \end{subfigure}
    \caption{Theoretical representation of thresholds from which choosing \gls*{di} over \gls*{ism} or \gls*{wmi}}
    \label{fig:direct_vs}
\end{figure*}

The results shown in \autoref{fig:direct_vs} are obtained by computing equations \eqref{eq:inversion_over_sherman} and \eqref{eq:inversion_over_woodbury} for all $k \in \llbracket 1, 4200 \rrbracket$ and $s \in \llbracket 1, 10000 \rrbracket$.

\vspace{0.5em}

\autoref{fig:direct_vs_sherman} shows that one must choose the \gls*{di} method over the \gls*{ism} method when $k > s / 2.4002 \approx 5s /12$.

\autoref{fig:direct_vs_woodbury} shows that one must choose the \gls*{di} method over the \gls*{wmi} method when $k > s / 3.7506$. We name this relation the empirical \gls*{di} over \gls*{wmi} threshold: 
\begin{equation}
    k > \frac{s}{3.7506}
    \label{eq:empirical_inversion_over_woodbury}
\end{equation}

\begin{table*}[pos=t]
    \centering
    \caption{Theoretical thresholds for choosing the \gls*{di} method over the \gls*{ism} or \gls*{wmi} methods}
    \begin{tabular}{c|c|c|c}
         & \makecell{\gls*{di} over \gls*{ism}} & \makecell{\gls*{di} over \gls*{wmi}} & Empirical \gls*{di} over \gls*{wmi} \\
         \hline
         $k$ & ${5 s^2} / ({12(s+1)})$ \eqref{eq:inversion_over_sherman} & $\frac{5}{6}k^3 + 4 sk^2 + $ $ 2 (s^2 - s) k - \frac{5}{6}s^3 > 0$ (\ref{eq:inversion_over_woodbury}, to solve with $s$ fixed) & $s / 3.7506$ \eqref{eq:empirical_inversion_over_woodbury} 
    \end{tabular}
    \label{table:theoretical_thresholds}
\end{table*}

\autoref{table:theoretical_thresholds} summarizes the different theoretical thresholds identified for choosing the \gls*{di} method over the \gls*{ism} or \gls*{wmi} methods.



\subsection{Experimental results}
\label{sec:experimental_results}
The following results were obtained on a laptop equipped with an Intel Core Ultra 5 135U CPU and 32 GB of RAM, running Windows 11. All results can be reproduced using the Python code and data available on GitHub\footnote{Link to the GitHub repository: \href{https://github.com/fgrivet/cost-trade-offs-matrix-inversion-update}{https://github.com/fgrivet/cost-trade-offs-matrix-inversion-update}}.

\vspace{0.5em}

We generate a random dataset of $S = 2000$ samples in $\mathbb{R}^{1287}$. This simulates 2000 vectors $v_n(\vecx)$ with $\vecx \in \mathbb{R}^{d=8}$ and polynomials of degree at most $n=5$: \\ $s =s_d(n) = \begin{pmatrix} d+n \\ n \end{pmatrix} = \begin{pmatrix} 8+5 \\ 5 \end{pmatrix} = 1287$.

\vspace{0.5em}

For every $k$ in $\{1, 2, 3, 4, 5, 10, 20, 30, 40, 50, 100, 200,$ $300, 400, 500, 750, 1000\}$, we compute the moment matrix $M_n(\mu_N)$ and its inverse $M_n(\mu_N)^{-1}$ with the first $N = S - k$ samples. Then we update the inverse of the moment matrix with the last $k$ samples using the three different methods and report the average execution times on the $ns=200$ simulations in \autoref{table:times_1287}.
Note that for the \gls*{ism} method, we reduced $ns$ to $50$ when $k > 50$.

\vspace{0.5em}

We also computed, for each method, the error of the inverse moment matrix: $e_m = \| I - M_{updated} \times M_{updated, m}^{-1} \|_F$, where $M_{updated, m}^{-1}$ corresponds to the updated inverse moment matrix $M_{updated}^{-1}$ found by the method $m \in \{ \text{DI, ISM,} \\ \text{WMI} \}$ and $\| . \|_F$ is the Frobenius norm \citep{frobenius} given by:
\begin{equation}
    \|A\|_F = \sqrt{\sum_{i, j} |a_{i, j}|^2}
    \label{eq:frobenius}
\end{equation}

\begin{table*}[pos=t]
    \centering
    \begin{minipage}{.4\textwidth}
        \centering
        \caption{Average execution time (in seconds) for a rank-$k$ \\ update of the moment matrix $M_n(\mu_N)$ of size $1287 \times 1287$ \\ \textbf{Smallest time} -- \underline{\textit{Second best time}}}
        \setlength{\tabcolsep}{5pt}
        \begin{tabular}{|c|c|c|c|}
            \hline
            \textbf{$k$} & \textbf{DI} & \textbf{ISM} & \textbf{WMI} \\\hline
            1 & 0.08594638 & \textbf{0.007743145} & \underline{\textit{0.0077865}} \\\hline
            \hline
            2 & 0.08421694 & \underline{\textit{0.01141714}} & \textbf{0.007551792} \\\hline
            3 & 0.08086838 & \underline{\textit{0.01424055}} & \textbf{0.005989846} \\\hline
            4 & 0.07941904 & \underline{\textit{0.02148158}} & \textbf{0.005903258} \\\hline
            5 & 0.07437832 & \underline{\textit{0.01883538}} & \textbf{0.004278867} \\\hline
            10 & 0.08314858 & \underline{\textit{0.06215603}} & \textbf{0.008458068} \\\hline
            \hline
            20 & \underline{\textit{0.08020028}} & 0.09959804 & \textbf{0.008605508} \\\hline
            30 & \underline{\textit{0.08923532}} & 0.1900465 & \textbf{0.01317246} \\\hline
            40 & \underline{\textit{0.08028041}} & 0.1966603 & \textbf{0.01141213} \\\hline
            50 & \underline{\textit{0.08450316}} & 0.2969778 & \textbf{0.01532174} \\\hline
            100 & \underline{\textit{0.08219219}} & 0.4851311 & \textbf{0.02341882} \\\hline
            200 & \underline{\textit{0.08573109}} & 0.9612609 & \textbf{0.04289328} \\\hline
            300 & \underline{\textit{0.09368211}} & 1.453023 & \textbf{0.06708244} \\\hline
            400 & \underline{\textit{0.09056586}} & 1.795042 & \textbf{0.09021928} \\\hline
            \hline
            500 & \textbf{0.09777304} & 2.207409 & \underline{\textit{0.1229474}} \\\hline
            750 & \textbf{0.1082857} & 3.867214 & \underline{\textit{0.2134477}} \\\hline
            1000 & \textbf{0.118939} & 3.829742 & \underline{\textit{0.3350951}} \\\hline
        \end{tabular}
        \label{table:times_1287}
    \end{minipage}
    \hfill
    \begin{minipage}{.5\textwidth}
    \centering
        \caption{Errors for a rank-$k$ update of the moment matrix $M_n(\mu_N)$ \\ of size $1287 \times 1287$ with conditioning $cond(M_n(\mu_N))$, $N=S-k$ \\ \textbf{Smallest error} -- \underline{\textit{Second best error}}}
        \setlength{\tabcolsep}{2pt}
        \begin{tabular}{|c|c|c|c||c|}
            \hline
            \textbf{$k$} & \textbf{DI} & \textbf{ISM} & \textbf{WMI} & \textbf{$cond(M_n(\mu_N))$} \\\hline
            1 & \textbf{1.580182e-13} & \underline{\textit{1.582999e-13}} & 1.583327e-13 & 83.9304 \\\hline
            \hline
            2 & \textbf{1.593879e-13} & 2.902351 & \underline{\textit{1.605027e-13}} & 84.0449 \\\hline
            3 & \textbf{1.594071e-13} & 4.358941 & \underline{\textit{1.595547e-13}} & 84.2276 \\\hline
            4 & \textbf{1.593055e-13} & 5.333518 & \underline{\textit{1.609318e-13}} & 85.1681 \\\hline
            5 & \textbf{1.59972e-13} & 6.150092 & \underline{\textit{1.604706e-13}} & 85.412 \\\hline
            10 & \textbf{1.598138e-13} & 9.544279 & \underline{\textit{1.610103e-13}} & 86.4186 \\\hline
            20 & \textbf{1.606954e-13} & 14.10532 & \underline{\textit{1.628296e-13}} & 88.7121 \\\hline
            30 & \textbf{1.607787e-13} & 17.76659 & \underline{\textit{1.641526e-13}} & 91.1043 \\\hline
            40 & \textbf{1.607618e-13} & 20.82128 & \underline{\textit{1.662739e-13}} & 93.5837 \\\hline
            50 & \textbf{1.61661e-13} & 23.75783 & \underline{\textit{1.677276e-13}} & 96.0594 \\\hline
            100 & \textbf{1.618503e-13} & 38.34813 & \underline{\textit{1.773381e-13}} & 104.694 \\\hline
            200 & \textbf{1.625134e-13} & 71.4727 & \underline{\textit{2.144143e-13}} & 136.435 \\\hline
            300 & \textbf{1.656789e-13} & 120.5727 & \underline{\textit{2.788295e-13}} & 219.31 \\\hline
            400 & \textbf{1.654148e-13} & 210.0006 & \underline{\textit{4.830752e-13}} & 325.742 \\\hline
            500 & \textbf{1.655513e-13} & 417.0248 & \underline{\textit{1.316689e-12}} & 604.421 \\\hline
            750 & \textbf{1.681712e-13} & 5.943877e+10 & \underline{\textit{21941.93}} & 3.44460e+18 \\\hline
            1000 & \textbf{1.694056e-13} & 2.559507e+11 & \underline{\textit{503624.6}} & 4.4279e+18 \\\hline
        \end{tabular}%
        \label{table:errors_1287}
    \end{minipage}
\end{table*}

The errors for each method and each value of $k$ are reported in \autoref{table:errors_1287}.
We observe that the errors associated with the WMI and ISM methods increase rapidly, when $k > 500$, indicating numerical instability. This behavior is largely explained by the poor conditioning $\left(\approx 10^{18}\right)$ of the empirical moment matrix constructed from limited data (1250 and 1000 samples, respectively).

\vspace{0.5em}

According to \citet{vershynin2018high}, and \citet{wainwright2019high}, increasing the number of samples improves conditioning  and numerical stability. So we repeated the experiment with a larger sample size: $S= 15000$. The errors for all methods stabilize around $10^{-13}$ across all tested values of $k$ (up to $k=1500$) and matrix sizes up to $s=2000$, confirming that the large errors observed in the small-sample regime are primarily due to ill conditioning of the moment matrix.

\vspace{0.5em}

For ISM method, the error still increases with $k$ for both $S = 2000$ and $S=15000$, reflecting the accumulation of floating-point rounding errors inherent to successive rank-one updates \cite{higham2002accuracy}. However, with sufficiently many samples, the error growth remains bounded and no longer exhibits the rapid instability observed for small sample sizes.

\vspace{0.5em}

The numerical stability of the Sherman-Morrison and Woodbury updates has been analyzed previously in \citep{stability}.

\subsection{Comparison with theoretical results}
The goal of this subsection is to validate the theoretical thresholds identified in \autoref{sec:theoretical_guide}, \autoref{table:theoretical_thresholds} and the values measured in \autoref{sec:experimental_results}, \autoref{table:times_1287}.

\vspace{0.5em}

\begin{table*}[pos=t]
    \centering
    \caption{Thresholds for choosing the \gls*{di} method over the \gls*{ism} or \gls*{wmi} methods with $s = 1287$: $d=8$ and $n=5$}
    \begin{tabular}{c|c|c|c}
         & \gls*{di} over \gls*{ism} & \gls*{di} over \gls*{wmi} & \makecell{Empirical \gls*{di} over \gls*{wmi}} \\
         \hline
         $k_{theoretical}$ &  $\eqref{eq:inversion_over_sherman} =\frac{5\times1287^2}{12(1287+1)} \approx 535.834$ & $\eqref{eq:inversion_over_woodbury} \approx 343.250$ & $\eqref{eq:empirical_inversion_over_woodbury} =\frac{1287}{3.7506} \approx 343.145$ \\
         \hline
         $k_{experimental}$ & Between $10$ and $20$ & \multicolumn{2}{c}{Between $400$ and $500$}
    \end{tabular}
    \label{table:theoretical_thresholds_1287}
\end{table*}

In \autoref{table:theoretical_thresholds_1287}, we present the theoretical and experimental thresholds for selecting the \gls*{di} method over the \gls*{ism} or \gls*{wmi} methods, with the matrix size $s$ fixed at 1287.
First, we observe that the empirical \gls*{di} over \gls*{wmi} approximation, given by equation \eqref{eq:empirical_inversion_over_woodbury} as $343.145$, is very close to the theoretical threshold of equation \eqref{eq:inversion_over_woodbury}, which is $343.250$. 
Second, as predicted in \autoref{sec:woodbury_over_sherman}, the \gls*{wmi} method proves to be more efficient than the \gls*{ism} method even for small values of $k$ ; specifically, starting from $k=2$.

\vspace{0.5em}

When comparing the thresholds for choosing the \gls*{di} method over the \gls*{ism} or \gls*{wmi} methods, we observe that the threshold \eqref{eq:inversion_over_sherman} for the \gls*{ism} method is significantly overestimated. The theoretical prediction is $535.834$, whereas experimentally, it falls between 10 and 20. Conversely, the threshold \eqref{eq:inversion_over_woodbury} for the \gls*{wmi} method is substantially underestimated. The theoretical prediction is $343.250$, but experimental results place it between 400 and 500, which is approximately $\frac{s}{3}$.
These discrepancies arise because our analysis considered only the number of operations (\gls*{flops}) and did not account for the efficiency of memory access (read/write operations) or the intrinsic Python optimizations for matrix calculations compared to vector calculations or iterative loops.

\subsection{Experimental guide for method selection based on new data volume \texorpdfstring{$k$}{k}}
\label{sec:experimental_guide}

We repeated the experiment conducted in \autoref{sec:experimental_results}. This time, we used $S=2000$ samples in $\mathbb{R}^s$ with $s$ ranging in $\{10, 20, 50, 100, 250, 500, 750, 1000\}$ and $ns=200$ simulations. However, for the \gls*{ism} method, when $s > 500$ and $k > 50$, we reduced $ns$ to $50$.

\begin{figure*}[pos=t]
    \centering
    \includegraphics[width=0.58\textwidth]{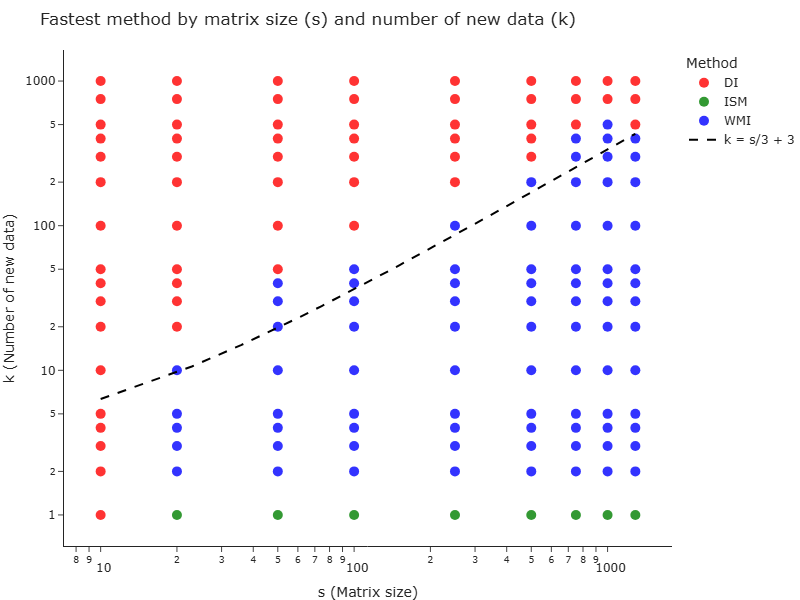}
    \caption{Fastest method for each (s, k) pair based on execution time}
    \label{fig:fastest_method_plot}
    \vspace{-0.5em}
\end{figure*}

\vspace{0.5em}

\autoref{fig:fastest_method_plot} illustrates the fastest method for performing the rank-$k$ update for each pair $(s, k)$. A green point signifies that \gls*{ism} (\autoref{alg:iterative_sherman_morrison}) is the fastest method, a blue point indicates that \gls*{wmi} (\autoref{alg:woodbury_matrix_identity}) is the fastest method, and a red point denotes that the \gls*{di} (\autoref{alg:direct_inversion}) is the fastest method.
Firstly, we notice that for a small matrix size ($s=10$), the \gls*{di} method outperforms the other two. Secondly, as expected, \gls*{ism} is faster for a rank-1 update. Then the \gls*{wmi} method is faster for a rank-$k$ update when $k < \frac{s}{3}$, and finally, for a large update, the \gls*{di} method is superior.

\vspace{0.5em}

Our recommendation, for a Python CPU implementation, is hence as follows:
\vspace{-0.5em}
\begin{itemize}
    \itemsep 0em
    \item Use the \gls*{ism} method (\autoref{alg:iterative_sherman_morrison}) for a rank-1 update.
    \item Use the \gls*{wmi} method (\autoref{alg:woodbury_matrix_identity}) for a rank-$k$ update when $k \leq \frac{s}{3}$, where $s$ is the matrix size.
    \item Use the \gls*{di} method (\autoref{alg:direct_inversion}) if $k > \frac{s}{3}$.
\end{itemize}


\section{Conclusion}\label{sec:conclusion}

This paper compares three methods for updating the inverse of a matrix after a rank-$k$ update, assuming the original inverse is known. Although motivated by Christoffel-function-based outlier detection, the analysis applies more broadly to problems involving rank-$k$ updates of \gls*{spd} invertible matrices. Theoretical thresholds for method selection are derived and summarized in \autoref{table:theoretical_thresholds}. They are validated empirically with comprehensive experiments that account for implementation effects.

\vspace{0.5em}

This note includes practical guidance for Python CPU implementations. 
With $s$ denoting the matrix dimension and $k$ the update rank, it recommends the following rule:
\vspace{-0.5em}
\begin{itemize}
    \item For $k = 1$, use \gls*{ism} (\autoref{alg:iterative_sherman_morrison}).
    \item For $k \leq \frac{s}{3}$, use \gls*{wmi} (\autoref{alg:woodbury_matrix_identity}).
    \item For $k > \frac{s}{3}$, use \gls*{di} (\autoref{alg:direct_inversion}).
\end{itemize}

\vspace{0.5em}

Although the theoretical computational costs derived in this work are expressed in terms of the matrix dimension $s$ and the update rank $k$, and are therefore independent of any specific programming language or hardware platform, the practical recommendations and proposed selection rule are empirically validated only for Python implementations executed on CPU. In particular, the quantitative thresholds identified in this study depend on memory-access patterns characteristic of standard Python numerical linear algebra libraries, and should not be assumed to transfer directly to other languages, libraries, or hardware architectures.

\vspace{0.5em}

Several directions for future work emerge from this study. First, it would be of interest to extend the empirical comparison to other programming languages and computational platforms, including compiled implementations such as C++ and GPU-accelerated environments, where differences in parallelism and memory hierarchies may significantly affect performance. Second, beyond inverse update strategies, a fundamental limitation of Christoffel-function-based anomaly detection lies in the size $s$ of the moment matrix, which grows rapidly with the ambient dimension. Future work could therefore focus on reducing the effective dimension of this matrix in order to lower computational cost and enable CF-based scoring to scale to high-dimensional data streams.

\printcredits

\section*{Data availability}
All code and datasets are available on GitHub: \\ \href{https://github.com/fgrivet/cost-trade-offs-matrix-inversion-update}{github.com/fgrivet/cost-trade-offs-matrix-inversion-update}. All the tests conducted in this paper can be reproduced. The README file explains how to use the code.

\section*{Declaration of generative AI and AI-assisted technologies in the manuscript preparation process}
During the preparation of this work the authors used Mistral in order to check the text grammar. After using this tool/service, the authors reviewed and edited the content as needed and hence take full responsibility for the content of the submitted article.

\section*{Declaration of competing interest}
The authors declare the following financial interests/personal relationships which may be considered as potential competing interests: Louise Travé-Massuyès reports financial support was provided by AI Interdisciplinary Institute ANITI. 
\section*{Acknowledgments}
This work has benefited from the AI Interdisciplinary Institute ANITI funded by the France 2030 program under the Grant agreements n°ANR-19-P3IA-0004 and n°ANR-23-IACL-0002.

The authors would like to thank Jean-Bernard Lasserre (LAAS-CNRS) and Didier Henrion (LAAS-CNRS) for their insightful discussions and contributions to the \acrlong{cf} method used in this paper.

\appendix
\section{Computational costs of generic elementary products}
\label{sec:computation_cost_generic}

In this appendix, we detail the computational costs of the intermediate steps employed in \autoref{sec:computational_costs}, expressed in terms of \gls*{flops}. 

\paragraph{Notations --} We adopt the following generic notation: scalars are denoted by lowercase letters (e.g., $a$), vectors by boldface lowercase letters (e.g., $\veca$), and matrices by uppercase letters (e.g., $A$). The dimensions of vectors and matrices are represented by $p$, $q$ and $m$. The computational costs are denoted by $C^i_j$, where the superscript $i$ indicates the dimension of the left element in the multiplication, and the subscript $j$ indicates the dimension of the right element.


\subsection{Computational cost of multiplying a row-vector by a column-vector}

\begin{figure}[pos=h]
    \centering
    \resizebox{!}{7em}{

\begin{tikzpicture}[
    mymatrix/.style={
        matrix of math nodes,
        nodes in empty cells,
        nodes={minimum width=0.5cm, minimum height=0.5cm, outer sep=0pt},
        left delimiter={[},
        right delimiter={]},
        column sep=0pt,
        row sep=0pt
    },
    rednode/.style={fill=red!20},
    bluenode/.style={fill=blue!20}
]


\matrix[mymatrix] (M2) {
    |[rednode]| a_1 & |(n22)| \cdots & |[bluenode]| a_{p-1} & |(n42)|  a_p \\
};

\matrix[mymatrix, right=1cm of M2] (M1) {
    |[rednode]| b_1 \\
    |(n21)|  \vdots \\
    |[bluenode]|   b_{p-1} \\
    |(n41)|  b_p \\
};

\node[right=1cm of M1] (R) {$w$};

\node[right=0.3cm of M2] (times) {$\times$};
\node[left=0.3cm of R] (equals) {$=$};

\end{tikzpicture}}
    \caption{Representation of the product of a row-vector by a column-vector}
    \label{fig:rvec_cvec}
\end{figure}
\vspace{-1em}

Let $\veca \in \mathbb{R}^{1 \times p}$, $\vecb \in \mathbb{R}^{p \times 1}$ and $w \in \mathbb{R}$ such that $w = \veca \times \vecb$ (\autoref{fig:rvec_cvec}).

$w = \overset{p}{\underset{i=1}{\sum}} a_i \times b_i$, i.e. $p$ multiplications and $p-1$ additions, which provides the following cost:
\begin{equation}
    \mathcal{C}^{1 \times p}_{p \times 1}= 2p - 1 ~ \gls*{flops}
    \label{eq:produit_vecteur_l_vecteur_c}.
\end{equation}

\subsection{Computational cost of multiplying a column-vector by a row-vector}

\begin{figure}[pos=h]
    \centering
    \resizebox{\linewidth}{!}{

\begin{tikzpicture}[
    mymatrix/.style={
        matrix of math nodes,
        nodes in empty cells,
        nodes={minimum width=0.3cm, minimum height=0.3cm, outer sep=0pt},
        left delimiter={[},
        right delimiter={]},
        column sep=0pt,
        row sep=0pt
    },
    rednode/.style={fill=red!20},
    bluenode/.style={fill=blue!20}
]

\matrix[mymatrix] (M1) {
    |[rednode]| a_1 \\
    |(n21)|  a_2 \\
    |(n31)|   \vdots \\
    |[bluenode]|  a_p \\
};

\matrix[mymatrix, above right=0pt and 1cm of M1.north east, anchor= north west] (M2) {
    |[rednode]| b_1 & |(n22)| \cdots & |[bluenode]| b_{q-1} & |(n42)|  b_q \\
};

\matrix[mymatrix, below right=0pt and 1cm of M2.north east, anchor=north west] (R) {
    |[rednode]| W_{1, 1} & |(r12)| & |(r13)| & |(r14)| \\
    |(r21)| & |(r22)| & |(r23)| & |(r24)| \\
    |(r31)| & |(r32)| & |(r33)| & |(r34)| \\
    |(r41)| & |(r42)| & |[bluenode]| W_{p, q-1}& |(r44)| \\
};

\node[right=0.3cm of M1] (times) {$\times$};
\node[left=0.3cm of R] (equals) {$=$};

\end{tikzpicture}}
    \vspace{-1em}
    \caption{Representation of the product of a column-vector by a row-vector}
    \label{fig:cvec_rvec}
\end{figure}
\vspace{-1em}

Let $\veca \in \mathbb{R}^{p \times 1}$, $\vecb \in \mathbb{R}^{1 \times q}$ and $W \in \mathbb{R}^{p \times q}$ such that $W = \veca \times \vecb$ (\autoref{fig:cvec_rvec}).

For every element of $W$, we have $W_{i, j} = a_i \times b_j$, i.e. 1 multiplication for $p \times q$ elements, which provides the following cost: 
\begin{equation}
    \mathcal{C}^{p \times 1}_{1 \times q}= p \times q ~\gls*{flops}
    \label{eq:produit_vecteur_c_vecteur_l}.
\end{equation}
When considering two vectors of equal length $p$, the computational cost of their product is $\mathcal{C}^{p \times 1}_{1 \times p}=p^2 ~ \gls*{flops}$.


\subsection{Computational cost of multiplying a row-vector by a matrix}

\begin{figure}[pos=h]
    \centering
    \resizebox{\linewidth}{!}{

\begin{tikzpicture}[
    mymatrix/.style={
        matrix of math nodes,
        nodes in empty cells,
        nodes={minimum width=0.3cm, minimum height=0.3cm, outer sep=0pt},
        left delimiter={[},
        right delimiter={]},
        column sep=0pt,
        row sep=0pt
    },
    rednode/.style={fill=red!20},
    bluenode/.style={fill=blue!20},
    purplenode/.style={fill=purple!20},
]

\matrix[mymatrix] (M1) {
    |(a11)| a_1 &
    |(a21)| a_2 &
    |(a31)| \cdots &
    |(a41)| a_p \\
};

\matrix[mymatrix, right=1 cm of M1] (R) {
    |(r11)| B_{1, 1} & |(r12)| & |(r13)| & |(r14)| B_{1, q}\\
    |(r21)| \vdots & |(r22)| & |(r23)| & |(r24)| \vdots \\
    |(r31)| \vdots & |(r32)| & |(r33)| & |(r34)| \vdots \\
    |(r41)| B_{p, 1} & |(r42)| & |(r43)| & |(r44)| B_{p, q} \\
};

\matrix[mymatrix, right=1 cm of R] (M2) {
    |[rednode]| w_1 &
    |(n21)|  w_2 &
    |(n31)|   \cdots &
    |[bluenode]|  w_q \\
};

\begin{scope}[on background layer]
    \node[fit=(r11) (r41), fill=red!20, inner sep=0pt, draw=none] {};
    \node[fit=(r14) (r44), fill=blue!20, inner sep=0pt, draw=none] {};
    \node[fit=(a11) (a41), fill=purple!20, inner sep=0pt, draw=none] {};
\end{scope}

\node[right=0.3cm of M1] (times) {$\times$};
\node[right=0.3cm of R] (equals) {$=$};

\end{tikzpicture}}
    \vspace{-1em}
    \caption{Representation of the product of a row-vector by a matrix}
    \label{fig:rvec_mat}
\end{figure}
\vspace{-1em}

Let $\veca \in \mathbb{R}^{1 \times p}$, $B \in \mathbb{R}^{p \times q}$ and $\vecw \in \mathbb{R}^{1 \times q}$ such that $\vecw = \veca \times B$ (\autoref{fig:rvec_mat}).

For every element of $\vecw$, we have $w_j = \overset{p}{\underset{i=1}{\sum}} a_i \times B_{i, j}$, i.e. $p$ multiplications and $p-1$ additions for $q$ elements, which equals to $q \times (2p - 1)$, which provides the following cost:
\begin{equation}
    \mathcal{C}^{1 \times p}_{p \times q}= 2 pq - q ~\gls*{flops}.
    \label{eq:produit_vecteur_l_matrice}
\end{equation}

When considering a square matrix of size $p \times p$, the computational cost of this product is $\mathcal{C}^{1 \times p}_{p \times p}=2 p^2 - p ~\gls*{flops}$.

\subsection{Computational cost of multiplying a matrix by a column-vector}

\begin{figure}[pos=h]
    \centering
    \resizebox{!}{7em}{

\begin{tikzpicture}[
    mymatrix/.style={
        matrix of math nodes,
        nodes in empty cells,
        nodes={minimum width=0.3cm, minimum height=0.3cm, outer sep=0pt},
        left delimiter={[},
        right delimiter={]},
        column sep=0pt,
        row sep=0pt
    },
    rednode/.style={fill=red!20},
    bluenode/.style={fill=blue!20},
    purplenode/.style={fill=purple!20},
]

\matrix[mymatrix] (R) {
    |(r11)| A_{1, 1} & |(r12)| \cdots & |(r13)| \cdots & |(r14)| A_{1, q}\\
    |(r21)| & |(r22)| & |(r23)| & |(r24)| \\
    |(r31)| & |(r32)| & |(r33)| & |(r34)| \\
    |(r41)| A_{p, 1} & |(r42)| \cdots & |(r43)| \cdots & |(r44)| A_{p, q} \\
};

\matrix[mymatrix, right=1 cm of R] (M1) {
    |(a11)| b_1 \\
    |(a21)| b_2 \\
    |(a31)| \vdots \\
    |(a41)| b_q \\
};

\matrix[mymatrix, right=1 cm of M1] (M2) {
    |[rednode]| w_1 \\
    |(n21)|  w_2 \\
    |(n31)|   \vdots \\
    |[bluenode]|  w_p \\
};

\begin{scope}[on background layer]
    \node[fit=(r11) (r14), fill=red!20, inner sep=0pt, draw=none] {};
    \node[fit=(r41) (r44), fill=blue!20, inner sep=0pt, draw=none] {};
    \node[fit=(a11) (a41), fill=purple!20, inner sep=0pt, draw=none] {};
\end{scope}

\node[right=0.3cm of R] (times) {$\times$};
\node[right=0.3cm of M1] (equals) {$=$};

\end{tikzpicture}}
    \caption{Representation of the product of a matrix by a column-vector}
    \label{fig:mat_cvec}
\end{figure}

Let $A \in \mathbb{R}^{p \times q}$, $\vecb \in \mathbb{R}^{q \times 1}$ and $\vecw \in \mathbb{R}^{p \times 1}$ such that $\vecw = A \times \vecb$ (\autoref{fig:mat_cvec}).

For every element of $\vecw$, we have $w_j = \overset{q}{\underset{i=1}{\sum}} A_{j, i} \times b_i$, i.e. $q$ multiplications and $q-1$ additions for $p$ elements, which equals to $p \times (2q - 1)$, which provides the following cost:
\begin{equation}
    \mathcal{C}^{p \times q}_{q \times 1}= 2 pq - p ~\gls*{flops}
    \label{eq:produit_matrice_vecteur_c}.
\end{equation}

When considering a square matrix of size $p \times p$, the computational cost of this product is $\mathcal{C}^{p \times p}_{p \times 1}=2 p^2 - p ~\gls*{flops}$

\subsection{Computational cost of multiplying a matrix by a matrix}

\begin{figure}[pos=h]
    \centering
    \resizebox{\linewidth}{!}{

\begin{tikzpicture}[
    mymatrix/.style={
        matrix of math nodes,
        nodes in empty cells,
        nodes={minimum width=0.25cm, minimum height=0.25cm, outer sep=0pt},
        left delimiter={[},
        right delimiter={]},
        column sep=0pt,
        row sep=0pt
    },
    rednode/.style={fill=red!20},
    bluenode/.style={fill=blue!20},
    purplenode/.style={fill=purple!20},
]

\matrix[mymatrix] (A) {
    |(a11)| A_{1, 1} & |(a13)| \cdots & |(a14)| A_{1, m}\\
    |(a21)| & |(a23)| & |(a24)| \\
    |(a31)| & |(a33)| & |(a34)| \\
    |(a41)| A_{p, 1} & |(a43)| \cdots & |(a44)| A_{p, m} \\
};

\matrix[mymatrix, right=0.7cm of A] (B) {
    |(b11)| B_{1, 1} & |(b12)| B_{1, 2} & |(b13)| \cdots & |(b14)| B_{1, q}\\
    |(b21)| \vdots & |(b22)| \vdots & |(b23)| & |(b24)| \vdots \\
    |(b31)| \vdots & |(b32)| \vdots & |(b33)| & |(b34)| \vdots \\
    |(b41)| B_{m, 1} & |(b42)| B_{m, 2} & |(b43)| \cdots & |(b44)| B_{m, q} \\
};

\matrix[mymatrix, right=0.7cm of B] (R) {
    |[rednode]| W_{1, 1} & |(r12)| \cdots & |(r13)| \cdots & |(r14)| W_{1, q}\\
    |(r21)| & |(r22)| & |(r23)| & |(r24)| \\
    |(r31)| & |(r32)| & |(r33)| & |(r34)| \\
    |(r41)| W_{p, 1} & |[bluenode]| W_{p, 2} & |(r43)| \cdots & |(r44)| W_{p, q} \\
};

\begin{scope}[on background layer]
    \node[fit=(a11) (a14), fill=red!20, inner sep=0pt, draw=none] {};
    \node[fit=(b11) (b41), fill=red!20, inner sep=0pt, draw=none] {};
    \node[fit=(a41) (a44), fill=blue!20, inner sep=0pt, draw=none] {};
    \node[fit=(b12) (b42), fill=blue!20, inner sep=0pt, draw=none] {};
\end{scope}

\node[right=0.15cm of A] (times) {$\times$};
\node[right=0.15cm of B] (equals) {$=$};

\end{tikzpicture}}
    \vspace{-1.8em}
    \caption{Representation of the product of a matrix by a matrix}
    \label{fig:mat_mat}
\end{figure}
\vspace{-1em}

Let $A \in \mathbb{R}^{p \times m}$, $B \in \mathbb{R}^{m \times q}$ and $W \in \mathbb{R}^{p \times q}$ such that $W = A \times B$ (\autoref{fig:mat_mat}).

For every element of $W$, we have $W_{i, j} = \overset{m}{\underset{l=1}{\sum}} A_{i, l} \times B_{l, j}$, i.e. $m$ multiplications and $m-1$ additions for $p \times q$ elements, which equals to $pq \times (2m - 1)$, which provides the following cost:
\begin{equation}
    \mathcal{C}^{p \times m}_{m \times q}= 2pqm - pq ~ \gls*{flops}.
    \label{eq:produit_matrice_matrice}
\end{equation}

When considering squares matrices of size $p \times p$, the computational cost of this product is $\mathcal{C}^{p \times p}_{p \times p}= 2 p^3 - p^2 ~\gls*{flops}$.

\bibliography{bib}
\bibliographystyle{cas-model2-names}

\end{document}